\documentclass[runningheads]{llncs}
\usepackage{IEEEtrantools}
\usepackage{graphicx}
\usepackage{amsmath}
\usepackage{amssymb}
\usepackage{gensymb}
\usepackage{xfrac}
\usepackage{algorithmic}
\usepackage{array}
\usepackage{url}
\usepackage{comment}
\usepackage{xcolor}
\usepackage{setspace}
\usepackage{bm}
\usepackage{multirow}
\usepackage[normalem]{ulem}
\usepackage{pifont}
\usepackage{acronym}
\usepackage{hyperref}

\acrodef{NN}{neural network}
\acrodef{CNN}{convolutional neural network}
\acrodef{MLP}{multilayer perceptron}
\acrodef{NAS}{Neural Architecture Search}
\acrodef{BN}{batch normalization}
\acrodef{BO}{Bayesian Optimization}
\acrodef{FMNIST}{Fashion MNIST}
\acrodef{config}{configuration}
\acrodef{conv}{convolutional}
\acrodef{DnC}{Deep-n-Cheap}

\newcommand{\eg} {\emph{e.g.,~}}
\newcommand{\ie} {\emph{i.e.,~}}
\newcommand{\ttr} {t_{\text{tr}}}

\begin{document}

\title{Deep-n-Cheap: An Automated Search Framework for Low Complexity Deep Learning\thanks{Supported by Defense Threat Reduction Agency (DTRA), USA.}
}

\titlerunning{Deep-n-Cheap}

\author{Sourya Dey \and
Saikrishna C. Kanala \and
Keith M. Chugg \and
Peter A. Beerel
}
\authorrunning{S. Dey et al.}
%
\institute{University of Southern California, Los Angeles CA 90089, USA
\email{\{souryade,kanala,chugg,pabeerel\}@usc.edu}}

\maketitle  

\begin{abstract}
We present Deep-n-Cheap -- an open-source AutoML framework to search for deep learning models. This search includes both architecture and training hyperparameters, and supports convolutional neural networks and multi-layer perceptrons. Our framework is targeted for deployment on both benchmark and custom datasets, and as a result, offers a greater degree of search space customizability as compared to a more limited search over only pre-existing models from literature. We also introduce the technique of `search transfer', which demonstrates the generalization capabilities of the models found by our framework to multiple datasets.

Deep-n-Cheap includes a user-customizable complexity penalty which trades off performance with training time or number of parameters. Specifically, our framework results in models offering performance comparable to state-of-the-art while taking 1-2 orders of magnitude less time to train than models from other AutoML and model search frameworks. Additionally, this work investigates and develops various insights regarding the search process. In particular, we show the superiority of a greedy strategy and justify our choice of Bayesian optimization as the primary search methodology over random / grid search.

\end{abstract}

\section{Introduction}\label{sec-intro}
Artificial \acp{NN} in deep learning systems are critical drivers of emerging technologies such as computer vision, text classification, and autonomous applications. In particular, \acp{CNN} are used for image related tasks while \acp{MLP} can be used for general purpose classification tasks. Manually designing these \acp{NN} is challenging since they typically have a large number of interconnected layers \cite{Krizhevsky2012_alexnet,Komodakis2016_WRN} and require a large number of decisions to be made regarding \emph{hyperparameters}. These hyperparameters, as opposed to trainable parameters like weights and biases, are not learned by the network. They need to be specified and adjusted by an external entity, \ie the designer. They can be broadly grouped into two categories -- a) architectural hyperparameters, such as the type of each layer and the number of nodes in it, and b) training hyperparameters, such as the learning rate and batch size. The difficulty of manually designing hyperparameters to find a good \ac{NN} is exacerbated by the fact that several hyperparameters interact with each other to have a combined effect on the final performance. 

\subsubsection{Motivation and Related Work:}
The problem of searching for good \acp{NN} has resulted in several efforts towards automating this process. These efforts include \emph{AutoML frameworks} such as Auto-Keras \cite{Jin2019_autokeras}, AutoGluon \cite{autogluon} and Auto-PyTorch \cite{Mendoza2018_autopytorch}, which are open source software packages applicable to a variety of tasks and types of \acp{NN}. The major focus of these efforts is on providing user-friendly toolkits to search for good hyperparameter values. 

Several other efforts place more emphasis on novel techniques for the search process. These can be broadly grouped into \ac{NAS} efforts such as \cite{Pham2018_ENAS,Liu2019_DARTS,Liu2018_PNAS,Baker2017_rl,Miikkulainen2019_evolution,Real2019_amoebanet,Xie2017_genetic,Tan2019_efficientnet,Cai2018_proxylessnas,He2018_AMC}, and efforts that place a larger emphasis on training hyperparameters over architecture \cite{Li2017_hyperband,Snoek2012_spearmint,Bergstra2013,Thornton2013_autoweka}. 
An alternate grouping is on the basis of search methodology -- a) reinforcement learning 
\cite{Pham2018_ENAS,Zoph2018_searchtransfer,Baker2017_rl}, b) evolution / genetic operations 
\cite{Miikkulainen2019_evolution,Real2019_amoebanet,Xie2017_genetic}, and c) \acl{BO} \cite{Kandasamy2018_NASBOT,Thornton2013_autoweka,Snoek2012_spearmint,Swersky2013}. Although the efforts described in this paragraph often come with publicly available software, they are typically not intended for general purpose use, \eg the code release for \cite{Cai2018_proxylessnas} only allows reproducing \acp{NN} on two datasets. This differentiates them from AutoML frameworks.

Deep \acp{NN} often suffer from \textbf{complexity} bottlenecks -- either in storage, quantified by the \emph{total number of trainable parameters} $N_p$, or computational, such as the number of FLOPs or the time taken to perform training and/or inference. Prior efforts on \ac{NN} search penalize inference complexity in specific ways -- latency in \cite{Cai2018_proxylessnas}, FLOPs in \cite{Tan2019_efficientnet}, and both in \cite{He2018_AMC}. However, inference complexity is significantly different from training since the latter includes backpropagation and parameter updates every batch
. For example, the resulting network for CIFAR-10 in \cite{Cai2018_proxylessnas} takes a minute to perform inference, but hours to train. 
Moreover, while there is considerable interest in popular benchmark datasets
, in most real-world applications deep learning models need to be trained on custom datasets for which readymade, pre-trained models do not exist \cite{Mayo2019_curemetrix,Baldi2014_physics,Santana2016_selfdrivingcar}. This leads to an increasing number of resource-constrained devices needing to perform training on the fly, \eg self-driving cars. 

The computing platform is also important, \eg 
changing batch size has a greater effect on training time per epoch on GPU than CPU. Therefore, calculating the FLOP count is not always an accurate measure of the time and resources expended in training a \ac{NN}. Some previous works have proposed pre-defined sparsity \cite{Dey2019_JETCAS,Dey2017_ICANN} and stochastic depth \cite{Huang2016_stochasticdepth} to reduce training time, while \cite{howtotrainresnet} focuses on finding the quickest training time to get to a certain level of performance. Note that these are all manual methods, not search frameworks.

\subsubsection{Overview and Contributions:}
This paper introduces \emph{\ac{DnC}} -- an open-source\footnote{The code and documentation are available at \url{https://github.com/souryadey/deep-n-cheap}} AutoML framework to search for deep learning models. We specifically target the training complexity bottleneck by including a penalty for \emph{training time per epoch} $\ttr$ in our search objective. The penalty coefficient can be varied by the user to obtain a family of networks trading off performance and complexity. Additionally, we also support storage complexity penalties for $N_p$.

\ac{DnC} searches for both architecture and training hyperparameters. While the architecture search derives some ideas from literature, we have striven to offer the user a considerable amount of customizability in specifying the search space. This is important for training on custom datasets which can have significantly different requirements than those associated with benchmark datasets.

\ac{DnC} primarily uses \acf{BO} and currently supports classification tasks using \acp{CNN} and \acp{MLP}. A notable aspect is \emph{search transfer}, where we found that the best \acp{NN} obtained from searching over one dataset give good performance on a different dataset. This helps to improve generalization in \acp{NN} -- such as on custom datasets -- instead of purely optimizing for specific problems.

The following are the key contributions of this paper:
\begin{enumerate}
    \item \textbf{Complexity}: To the best of our knowledge, \ac{DnC} is the only AutoML framework targeting training complexity reduction. We show results on several datasets on both GPU and CPU. Our models achieve performance comparable to state-of-the-art, with training times that are 1-2 orders of magnitude less than those for models obtained from other AutoML and search efforts.
    \item \textbf{Usability}: \ac{DnC} offers a highly customizable three-stage search interface for both architecture and training hyperparameters. As opposed to Auto-Keras and AutoGluon, our search includes a) batch size that affects training times, and b) architectures beyond pre-existing ones found in literature. As a result, our target users include those who want to train quickly on custom datasets. As an example, our framework achieves the highest performance and lowest training times on the custom Reuters RCV1 dataset \cite{Dey2019_JETCAS}. We also introduce \emph{search transfer} to explore generalization capabilities of architectures to multiple datasets under different training hyperparameter settings.
    \item \textbf{Insights}: We conduct investigations into the search process and draw several insights that will help guide a deeper understanding of \acp{NN} and search methodologies. We introduce a new similarity measure for \ac{BO} and a new distance function for \acp{NN}. We empirically justify the value of our greedy three-stage search approach over less greedy approaches, and the superiority of \ac{BO} over random and grid search.
\end{enumerate}


The paper is structured as follows –– Sec. \ref{sec-approach} outlines our search methodology, Sec. \ref{sec-expts} our experimental results, Sec. \ref{sec-investigations} includes additional investigations and insights, Sec. \ref{sec-comp} compares with related work, and Sec. \ref{sec-conc} concludes the paper.

\section{Our Approach}\label{sec-approach}
Given a dataset, our framework searches for \ac{NN} \acp{config} through sequential stages in multiple search spaces. Each \ac{config} is trained for the same number of epochs, \eg 100. There have been works on extrapolating \ac{NN} performance from limited training \cite{Baker2017_perfpredict,Liu2018_PNAS}, however we train for a large number of epochs to predict with significant confidence the final performance of a \ac{NN} after convergence. Configs are mapped to objective values using:
\begin{IEEEeqnarray}{c}\label{eq-search_objective}
f(\text{Config}) = f_p(\text{Config}) + w_cf_c(\text{Config})
\end{IEEEeqnarray}
where $w_c$ controls the importance given to the complexity term. The goal of the search is to minimize $f$. Its components are:
\begin{IEEEeqnarray}{rCl}
f_p &=& 1 - \text{Best Validation Accuracy} \IEEEyesnumber \IEEEyessubnumber \label{eq-fp} \\
f_c &=& \frac{c}{c_0} \IEEEnonumber \IEEEyessubnumber \label{eq-fc}
\end{IEEEeqnarray}
where $c$ is the complexity metric for the current \ac{config} (either $\ttr$ or $N_p$), and $c_0$ is a reference value for the same metric (typically obtained for a high complexity \ac{config} in the space). Lower values of $w_c$ focus more on performance, \ie improving accuracy. One key contribution of this work is characterizing higher values of $w_c$ that lead to reduced complexity \acp{NN} that train fast -- these also reduce the search cost by speeding up the overall search process.

\subsection{Three-stage search process}\label{sec-approach-stages}

\begin{figure}[!t]
    \centering
    \includegraphics[width=\textwidth]{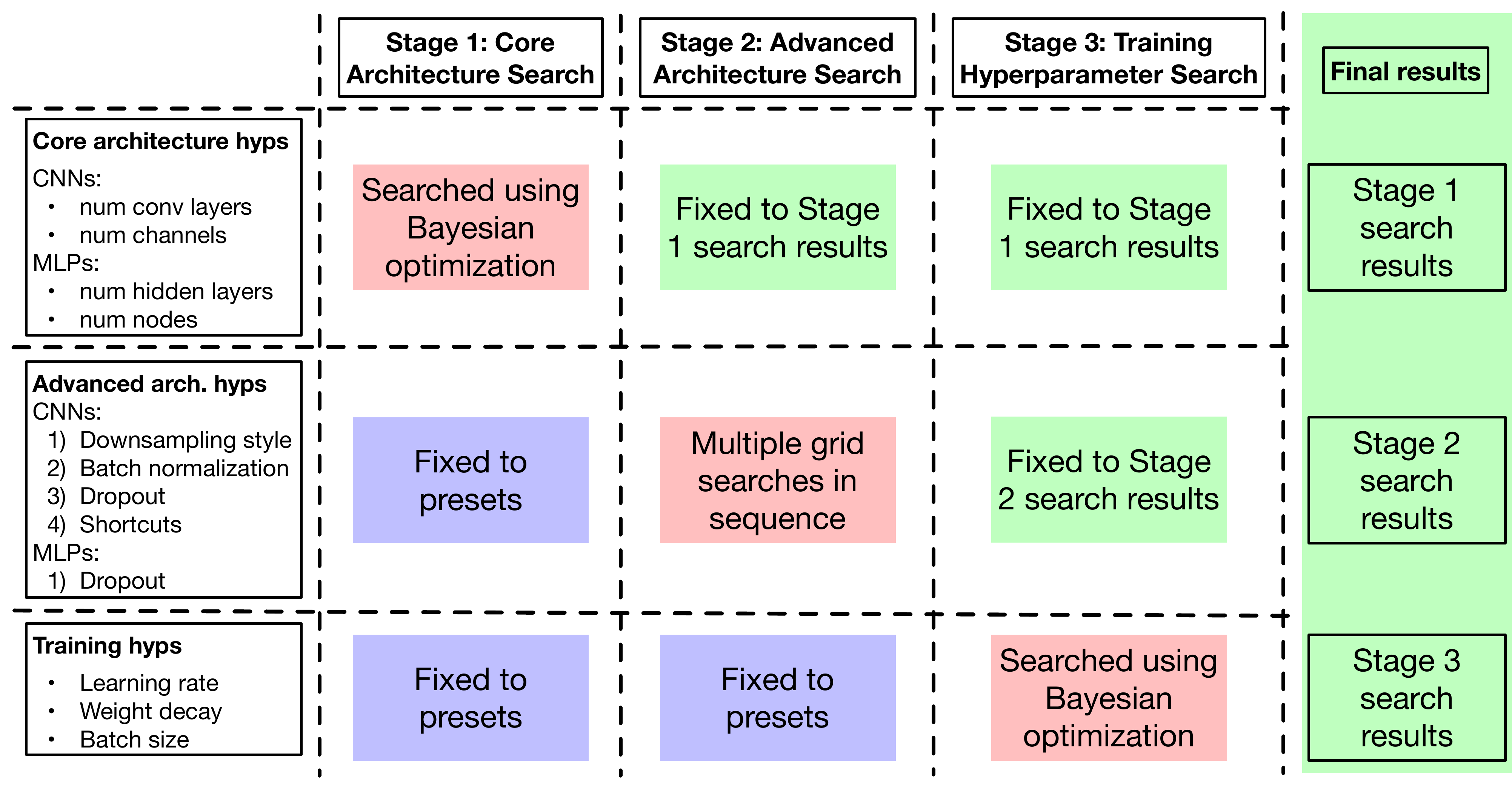}
    \caption{Three-stage search process for \ac{DnC}.}
    \label{fig-search_stages}
\end{figure}

\subsubsection{Stage 1 -- Core architecture search:}
For \acp{CNN}, the combined search space consists of the number of \ac{conv} layers and number of channels in each, 
while for \acp{MLP}, it is the number of hidden layers and number of nodes in each. Other architectural hyperparameters such as \ac{BN} and dropout layers and all training hyperparameters are fixed to presets that we found to work well across a variety of datasets and network depths. \ac{BO} is used to minimize $f$ and the corresponding best \ac{config} is the Stage 1 result.

\subsubsection{Stage 2 -- Advanced architecture search:}
This stage starts from the resulting architecture from Stage 1 and uses grid search to search for the following \ac{CNN} hyperparameters through a sequence of sub-stages -- 1) whether to use strides or max pooling layers for downsampling, 2) amount of \ac{BN} layers, 3) amount of dropout layers and drop probabilities, and 4) amount of shortcut connections. This is not a combined space, instead grid search first picks the downsampling choice leading to the minimum $f$ value, then freezes that and searches over \ac{BN}, and so on. This ordering yielded good empirical results, however, reordering is supported by the framework. For \acp{MLP}, there is a single grid search for dropout probabilities. As in the previous stage, training hyperparameters are fixed to presets. The result from Stage 2 is the result from the final sub-stage. 
    
\subsubsection{Stage 3 -- Training hyperparameter search:}
The architecture is finalized after Stage 2. In Stage 3 -- identical for \acp{CNN} and \acp{MLP} -- we search over the combined space of initial learning rate $\eta$, weight decay $\lambda$ and batch size, using \ac{BO} to minimize $f$. 
The final \ac{config} after Stage 3 comprises both architecture and training hyperparameters. The complete process is summarized in Fig. \ref{fig-search_stages}.

\subsection{Bayesian Optimization}\label{sec-approach-bayesopt}
\acl{BO} is useful for optimizing functions that are black-box and/or expensive to evaluate such as $f$, which requires \ac{NN} training. 
The initial step when performing \ac{BO} is to sample $n_1$ \acp{config} from the search space, $\{\bm{x}_1,\allowbreak \cdots,\allowbreak \bm{x}_{n_1}\}$, calculate their corresponding objective values, $\{f\left(\bm{x}_1\right),\allowbreak \cdots,\allowbreak f\left(\bm{x}_{n_1}\right)\}$, and form a Gaussian prior. The mean vector $\bm{\mu}$ is filled with the mean of the $f$ values, and covariance matrix $\bm{\Sigma}$ is such that $\Sigma_{ij} = \sigma\left(\bm{x}_i,\bm{x}_j\right)$, where $\sigma(\cdot,\cdot)$ is a \emph{kernel function} that takes a high value $\in[0,1]$ if \acp{config} $\bm{x}_i$ and $\bm{x}_j$ are similar.

Then the algorithm continues for $n_2$ steps, each step consisting of sampling $n_3$ \acp{config}, picking the \ac{config} with the maximum \emph{expected improvement}
, computing its $f$ value, and updating $\bm{\mu}$ and $\bm{\Sigma}$ accordingly. The reader is referred to \cite{Brochu2010_BOtutorial} for a complete tutorial on \ac{BO} -- where eq. (4) in particular has details of expected improvement. Note that \ac{BO} explores a total of $n_1+n_2n_3$ states in the search space, but the expensive $f$ computation only occurs for $n_1+n_2$ states.

\subsubsection{Similarity between \ac{NN} configurations:}
We begin by defining the \emph{distance} between values of a particular hyperparameter $k$ for two \acp{config} $\bm{x}_i$ and $\bm{x}_j$. Larger distances denote dissimilarity. We initially considered the distance functions defined in Sections 2 and 3 of \cite{Hutter2013}, 
but then adopted an alternate one that resulted in similar performance with less tuning. We call it the \emph{ramp} distance:
\begin{IEEEeqnarray}{c}
d\left(x_{ik},x_{jk}\right) = \omega_k \left(\frac{\left|x_{ik}-x_{jk}\right|}{u_k-l_k}\right)^{r_k} \IEEEyesnumber \IEEEyessubnumber \label{eq-distfn_ramp}
\end{IEEEeqnarray}
where $u_k$ and $l_k$ are respectively the upper and lower bounds for $k$, $\omega_k$ is a scaling coefficient, and $r_k$ is a fractional power used for stretching small differences. Note that $d$ is 0 when $x_{ik}=x_{jk}$, and reaches a maximum of $\omega_k$ when they are the furthest apart. $x_{ik}$ and $x_{jk}$ are computed in different ways depending on $k$:
\begin{itemize}
    \item If $k$ is batch size or number of layers, $x_{ik}$ and $x_{jk}$ are the actual values.
    \item If $k$ is $\eta$ or $\lambda$, $x_{ik}$ and $x_{jk}$ are the logarithms of the actual values.
    \item When $k$ is the hidden node configuration of a \ac{MLP}, we sum the nodes together across all hidden layers. This is because we found that the sum has a greater impact on $f$ than considering layers individually, \eg a \ac{config} with three 300-node hidden layers has a closer $f$ value to a \ac{config} with one 1000-node hidden layer than a \ac{config} with three 100-node hidden layers.
    \item When $k$ is the \ac{conv} channel configuration of a \ac{CNN}, we calculate individual distances for each layer. If the number of layers is different, the distance is maximum for each of the extra layers, \ie $\omega$. This idea is inspired from \cite{Hutter2013}, as compared to alternative similarity measures in \cite{Kandasamy2018_NASBOT,Jin2019_autokeras}. We follow this layer-by-layer comparison because our prior experiments showed that the representations learned by a certain \ac{conv} layer in a \ac{CNN} are similar to those learned by layers at the same depth in different \acp{CNN}. Additionally, this approach performed better than the summing across layers as in \acp{MLP}.
\end{itemize}

\begin{figure}[!t]
    \centering
    \includegraphics[width=0.85\textwidth]{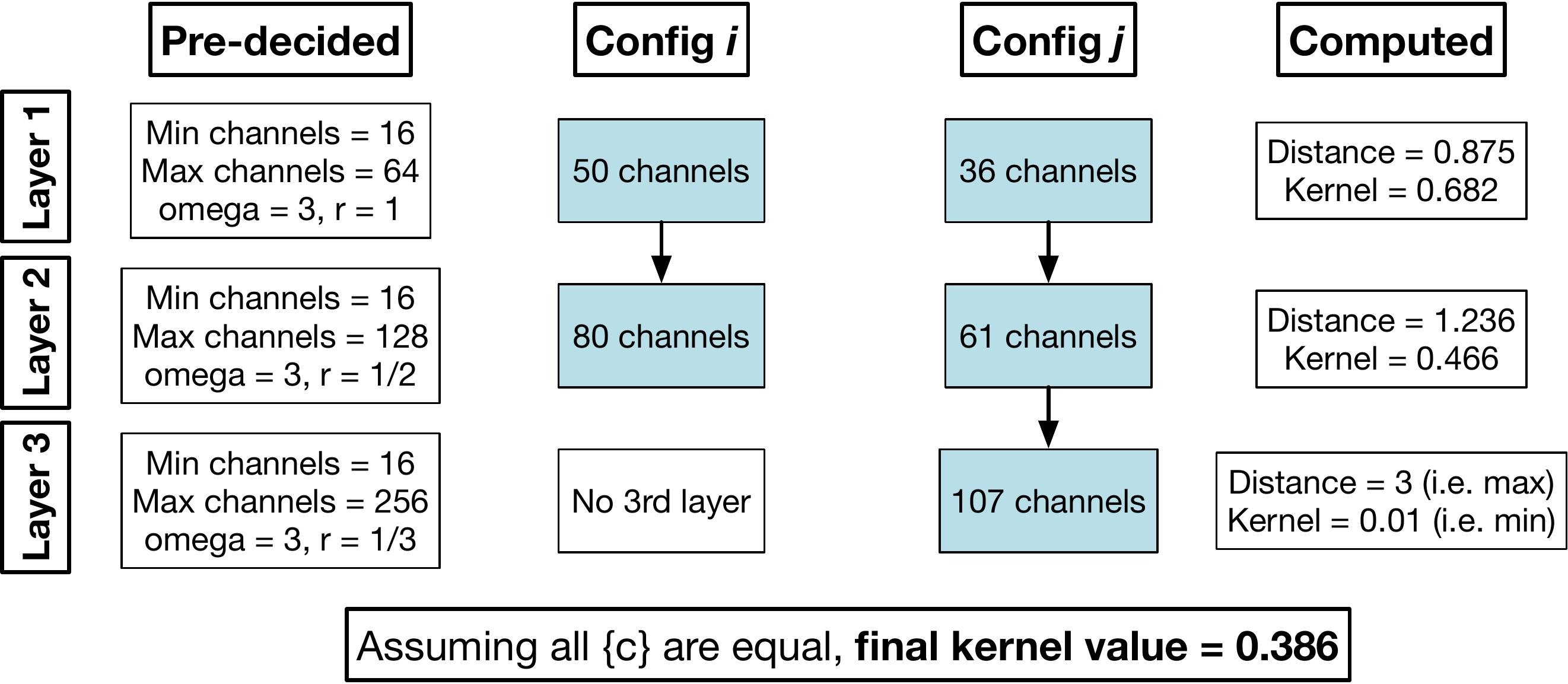}
    \caption{Calculating Stage 1 similarity for two \ac{conv} channel \acp{config}: $\bm{x}_i = [50,\allowbreak 80]$ and $\bm{x}_j = [36,\allowbreak 61,\allowbreak 107]$. Taking the 1st \ac{conv} layer as an example, the pre-decided values are $u_1=64$, $l_1=16$, $\omega_1=3$ and $r_1=1$ (more details on these choices in Sec. \ref{sec-expts}). The distance is $d_1 = 3\times\left[(50-36)/(64-16)\right]^1 = 0.875$, and kernel value is $\sigma_1 = \exp\left(-0.5\times0.875^2\right) = 0.682$. Similarly we get $\sigma_2=0.466$ and $\sigma_3=0.01$ (note that $d_3=\omega_3$ due to the absence of the 3rd layer in $\bm{x}_i$). Combining these using $s_1=s_2=s_3=\sfrac{1}{3}$ yields $\sigma\left(\bm{x}_i,\bm{x}_j\right) = 0.386$.}
    \label{fig-conv_distance}
\end{figure}

Each individual distance $d\left(x_{ik},x_{jk}\right)$ is converted to its kernel value $\sigma\left(x_{ik},x_{jk}\right)$ using the squared exponential function, then we take their convex combination for all $K$ hyperparameters using coefficients $\left\{s_k\right\}$ to finally get $\sigma\left(\bm{x}_i,\bm{x}_j\right)$. An example is given in Fig. \ref{fig-conv_distance}.
\begin{IEEEeqnarray}{rCl}
\sigma\left(x_{ik},x_{jk}\right) &=& \exp \left(-\frac{d^2(x_{ik}, x_{jk})}{2} \right) \IEEEnonumber \IEEEyessubnumber \label{eq-kernel_part} \\
\sigma\left(\bm{x}_i,\bm{x}_j\right) &=& \sum_{k=1}^{K}{s_k\sigma\left(x_{ik},x_{jk}\right)} \IEEEnonumber \IEEEyessubnumber \label{eq-kernel_whole}
\end{IEEEeqnarray}

\section{Experimental Results}\label{sec-expts}
This section presents results of our search framework on different datasets for both \ac{CNN} and \ac{MLP} classification problems, along with the search settings used. Note that most of these \emph{settings can be customized by the user} -- this leads to one of our key contributions of using limited knowledge from literature to enable wider exploration of \acp{NN} for various custom problems. We used the Pytorch library on two platforms: a) \emph{GPU} -- an Amazon Web Services p3.2xlarge instance that uses a single NVIDIA V100 GPU with 16 GB memory and 8 vCPUs, and b) \emph{CPU} -- a mid-2014 Macbook Pro CPU with 2.2 GHz Intel Core i7 processor and 16GB 1.6 GHz DDR3 RAM. For \ac{BO}, we used $n_1 = n_2 = 15$ and $n_3 = 1000$.

\subsection{\acp{CNN}}
All \ac{CNN} experiments are on GPU. The datasets used are CIFAR-10 and -100 with train-validation-test splits of 40k-10k-10k, and \ac{FMNIST} with 50k-10k-10k. Standard augmentation is always used -- channel-wise normalization, random crops from 4 pixel padding on each side, and random horizontal flips. Augmentation requires Pytorch data loaders that incur timing overheads, so we also show results on unaugmented CIFAR-10 where the whole dataset is loaded into memory at the beginning and $\ttr$ reduces as a result.

For \textbf{Stage 1}, we use \ac{BO} to search over \acp{CNN} with 4--16 \ac{conv} layers, the first of which has $c_1 \in \{16,\allowbreak 17,\allowbreak \cdots,\allowbreak 64\}$ channels and each subsequent layer has $c_{i+1} \in \{c_i,\allowbreak c_i+1,\allowbreak \cdots,\allowbreak \min\left(2c_i,512\right)\}$ channels. We allow the number of channels in a layer to have arbitrary integer values, not just fixed to multiples of 8. Kernel sizes are fixed to 3x3. Downsampling precedes layers where $c_i$ crosses 64, 128 and 256 (this is due to GPU memory limitations). During Stage 1, all \ac{conv} layers are followed by \ac{BN} and dropout with $30\%$ drop probability. Configs with more than 8 \ac{conv} layers have shortcut connections. Global average pooling and a softmax classifier follows the \ac{conv} portion. There are no hidden classifier layers since we empirically obtained no performance benefit. For both Stages 1 and 2, we used the default Adam optimizer with $\eta = 10^{-3}$, decayed by $80\%$ at the half and three-quarter points of training, batch size of 256, and $\lambda = \mathbb{I}(N_p\ge10^6)\times N_p/10^{11}$, $\mathbb{I}$ being the indicator function. We empirically found this rule to work well.

For \textbf{Stage 2}, the first grid search is over all possible combinations of using either strides or max pooling for the downsampling layers. Second, we vary the fraction of \ac{BN} layers through $\left[0,\sfrac{1}{4},\sfrac{1}{2},\sfrac{3}{4}\right]$. For example, if there are 7 \ac{conv} layers, a setting of $\sfrac{1}{2}$ will place \ac{BN} layers after \ac{conv} layers 2, 4, 6 and 7. Third, we vary the fraction of dropout layers in a manner similar to \ac{BN}, and drop probabilities over $[0.1,0.2]$ for the input layer and $[0.15,0.3,0.45]$ for all other layers. Finally, we search over shortcut connections -- none, every 4th layer, or every other layer. Note that any shortcut connection skips over 2 layers.

For \textbf{Stage 3}, we used \ac{BO} to search over a) $\eta\in\left\{10^x\right\}$ for $x \in [1,5]$, b) $\lambda\in\left\{10^x\right\}$ for $x \in [-6,-3]$, with $\lambda$ converted to $0$ when $x<-5$, and c) batch sizes in $[32,\allowbreak 33,\allowbreak \cdots,\allowbreak 512]$. We found that batch sizes that are not powers of 2 did not lead to any slowdown on the platforms used.

The penalty function $f_c$ uses normalized $\ttr$, since this is the major bottleneck in developing \acp{CNN}. Each \ac{config} was trained for 100 epochs on the train set and evaluated on the validation set to obtain $f_p$. We ran experiments for 5 different values of $w_c$: $[0,0.01,0.1,1,10]$. The best network from each search was then trained on the combined training and validation set, and evaluated on the test set for 300 epochs to get final test accuracies and $\ttr$ values.

\begin{figure}[!t]
    \centering
    \includegraphics[width=0.9\textwidth]{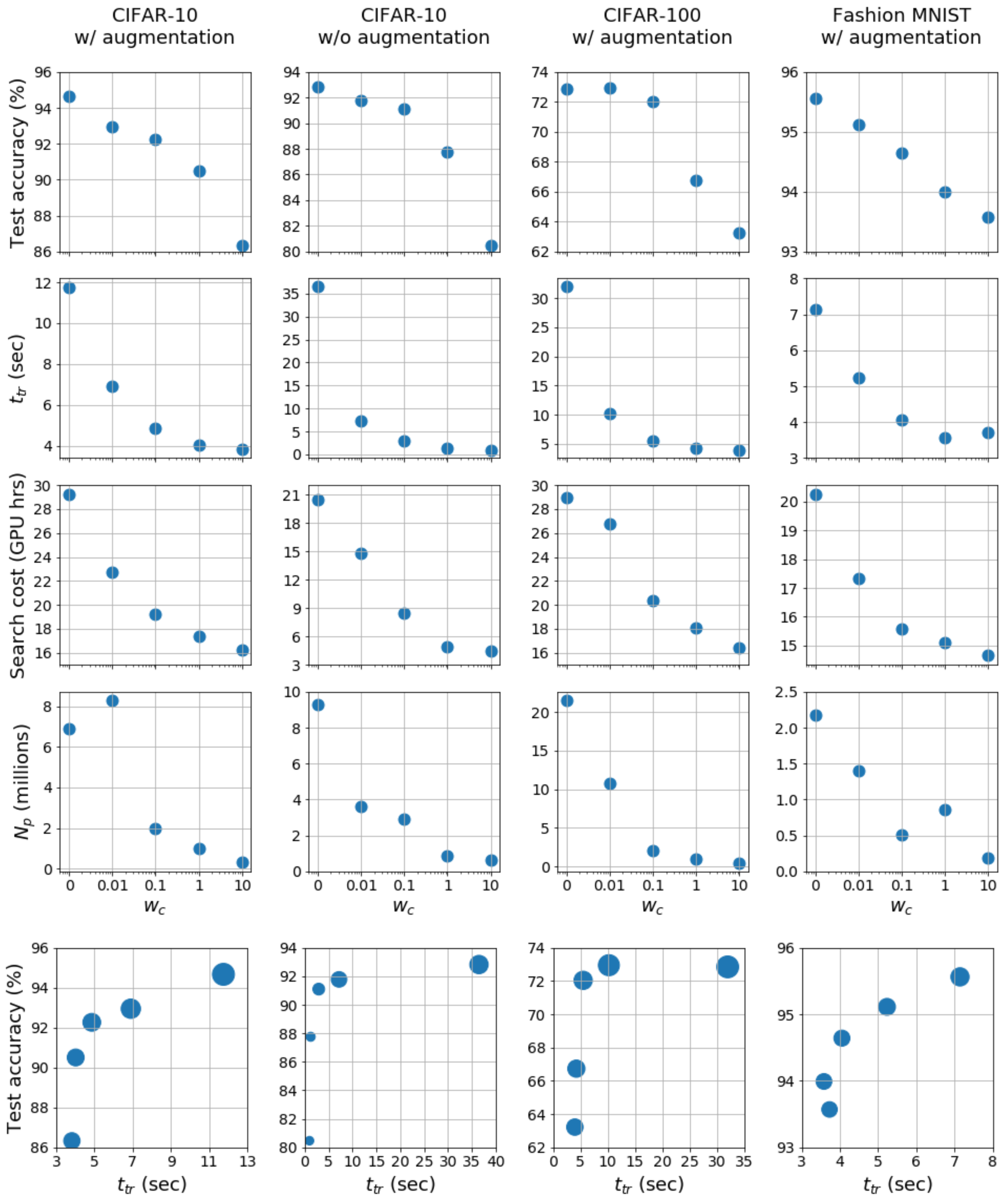}
    \caption{Characterizing a family of \acp{NN} for CIFAR-10 augmented (1st column), unaugmented (2nd column), CIFAR-100 augmented (3rd column) and \ac{FMNIST} augmented (4th column), obtained from \ac{DnC} for different $w_c$. We plot test accuracy in 300 epochs (1st row), $\ttr$ on combined train and validation sets (2nd row), search cost (3rd row) and $N_p$ (4th row), all against $w_c$. The 5th row shows the performance-complexity tradeoff, with dot size proportional to search cost.}
    \label{fig-cnn_results}
\end{figure}

As shown in Fig. \ref{fig-cnn_results}, we obtain a family of networks by varying $w_c$. Performance in the form of test accuracy trades off with complexity in the form of $\ttr$. The latter is correlated with search cost and $N_p$. The last row of figures directly plot the performance-complexity tradeoff. These curves rise sharply towards the left and flatten out towards the right, indicating diminishing performance returns as complexity is increased. This highlights one of our key contributions -- allowing the user to choose fast training \acp{NN} that perform well.

Taking augmented CIFAR-10 as an example, \ac{DnC} found the following best \ac{config} for $w_c=0$: 14 \ac{conv} layers with $\{c\} = (50,\allowbreak 52,\allowbreak 53,\allowbreak 59,\allowbreak 95,\allowbreak 96,\allowbreak 97,\allowbreak 120,\allowbreak 193,\allowbreak 239,\allowbreak 351,\allowbreak 385,\allowbreak 488,\allowbreak 496)$, the 4th layer has a stride of 2 while max pooling follows layers 8 and 10, \ac{BN} follows all \ac{conv} layers, dropout with drop probability $0.3$ follows every other \ac{conv} block, and skip connections are present for every other \ac{conv} block. The best found $\eta$ remains $10^{-3}$, batch size is 120 and $\lambda = 3.35\times10^{-5}$. We note that we achieve good performance with a \ac{NN} that has irregular $\{c\}$ values and is also not very deep -- the latter is consistent with the findings in \cite{Komodakis2016_WRN}. Also note that the best \ac{config} found for $w_c=10$ only has 4 \ac{conv} layers.

\subsection{\acp{MLP}}
We ran CPU experiments on the MNIST and \ac{FMNIST} datasets in permutation-invariant format (\ie images are flattened to a single layer of 784 input pixels) without any augmentation, and GPU experiments on the Reuters RCV1 dataset constructed as given in \cite{Dey2019_JETCAS}. Each dataset is loaded into memory in its entirety, eliminating data loader overheads.

For \textbf{Stage 1}, we search over 0--2 hidden layers for MNIST and \ac{FMNIST}, number of nodes in each being 20--400. These numbers change for RCV1 to 0--3 and 50--1000 since it is a larger dataset. Every layer is followed by a dropout layer with $20\%$ drop probability. Training hyperparameters are fixed as in the case of \acp{CNN}, with the difference that $\lambda = \mathbb{I}(N_p\ge10^4)\times N_p/10^9$ for MNIST and \ac{FMNIST} and $\lambda = \mathbb{I}(N_p\ge10^5)\times N_p/10^{10}$ for RCV1. For \textbf{Stage 2}, we do a grid search over drop probabilities in $[0,0.1,0.3,0.4,0.5]$, and for \textbf{Stage 3}, the training hyperparameter search is identical to \acp{CNN}.

\begin{figure}[!t]
    \centering
    \includegraphics[width=0.8\textwidth]{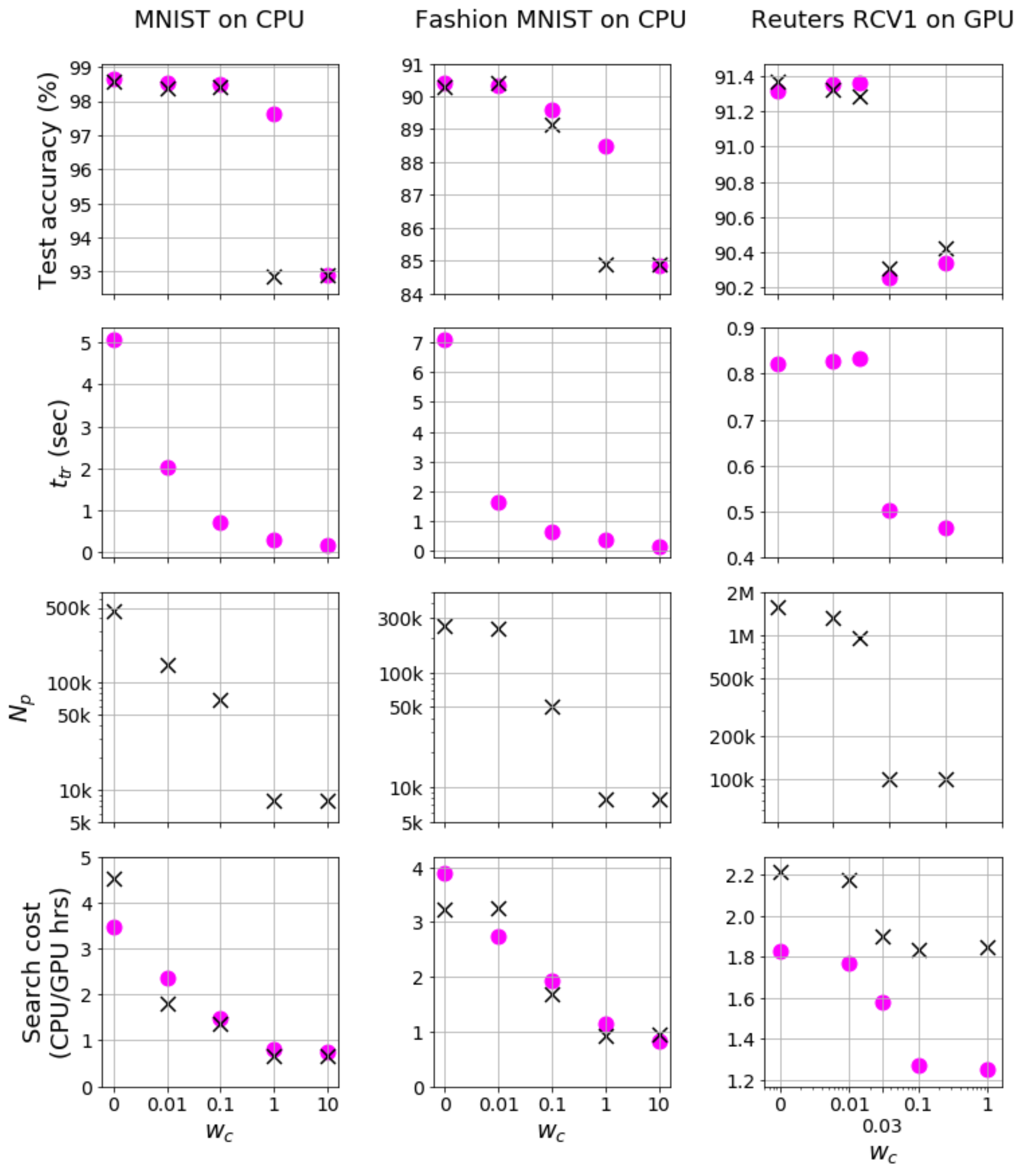}
    \caption{Characterizing a family of \acp{NN} for MNIST (1st column) and \ac{FMNIST} (2nd column) on CPU, and RCV1 (3rd column) on GPU, obtained from \ac{DnC} for different $w_c$. We plot test accuracy in 180 epochs (1st row), $\ttr$ on combined train and validation sets (2nd row), $N_p$ (3rd row), and search cost (4th row), all against $w_c$. The search penalty is $\ttr$ for the pink dots and $N_p$ for the black crosses.}
    \label{fig-mlp_results}
\end{figure}

We ran separate searches for individual penalty functions -- normalized $\ttr$ and normalized $N_p$. The latter is owing to the fact that \acp{MLP} often massively increase the number of parameters and thereby storage complexity of \acp{NN} \cite{Krizhevsky2012_alexnet}. The train-validation-test splits for MNIST and \ac{FMNIST} are 50k-10k-10k, and 178k-50k-100k for RCV1. Candidate networks were trained for 60 epochs and the final networks tested after 180 epochs. As before, $w_c \in [0,0.01,0.1,1,10]$ for MNIST and \ac{FMNIST}. For RCV1, the results for $w_c = 10$ were mostly similar to $w_c = 1$, so we replace $10$ with $0.03$. The plots against $w_c$ are shown in Fig. \ref{fig-mlp_results}, where pink dots are for $\ttr$ penalty and black crosses are for $N_p$ penalty.

The trends in Fig. \ref{fig-mlp_results} are qualitatively similar to those in Fig. \ref{fig-cnn_results}. When penalizing $N_p$, the two lowest complexity networks in each case have no hidden layers, so they both have exactly the same $N_p$ (results differ due to different training hyperparameters). Of interest is the subfigure on the bottom right, indicating much longer search times when penalizing $N_p$ as compared to $\ttr$. This is because time is not a factor when penalizing $N_p$, so the search picks smaller batch sizes that increase $\ttr$ with a view to improving performance. Interestingly enough, this does not actually lead to performance benefit as shown in the subfigure on the top-right, where the black crosses occupy similar locations as the pink dots.

\section{Investigations and insights}\label{sec-investigations}

\subsection{Search transfer}
\begin{figure}[!t]
    \centering
    \includegraphics[width=\textwidth]{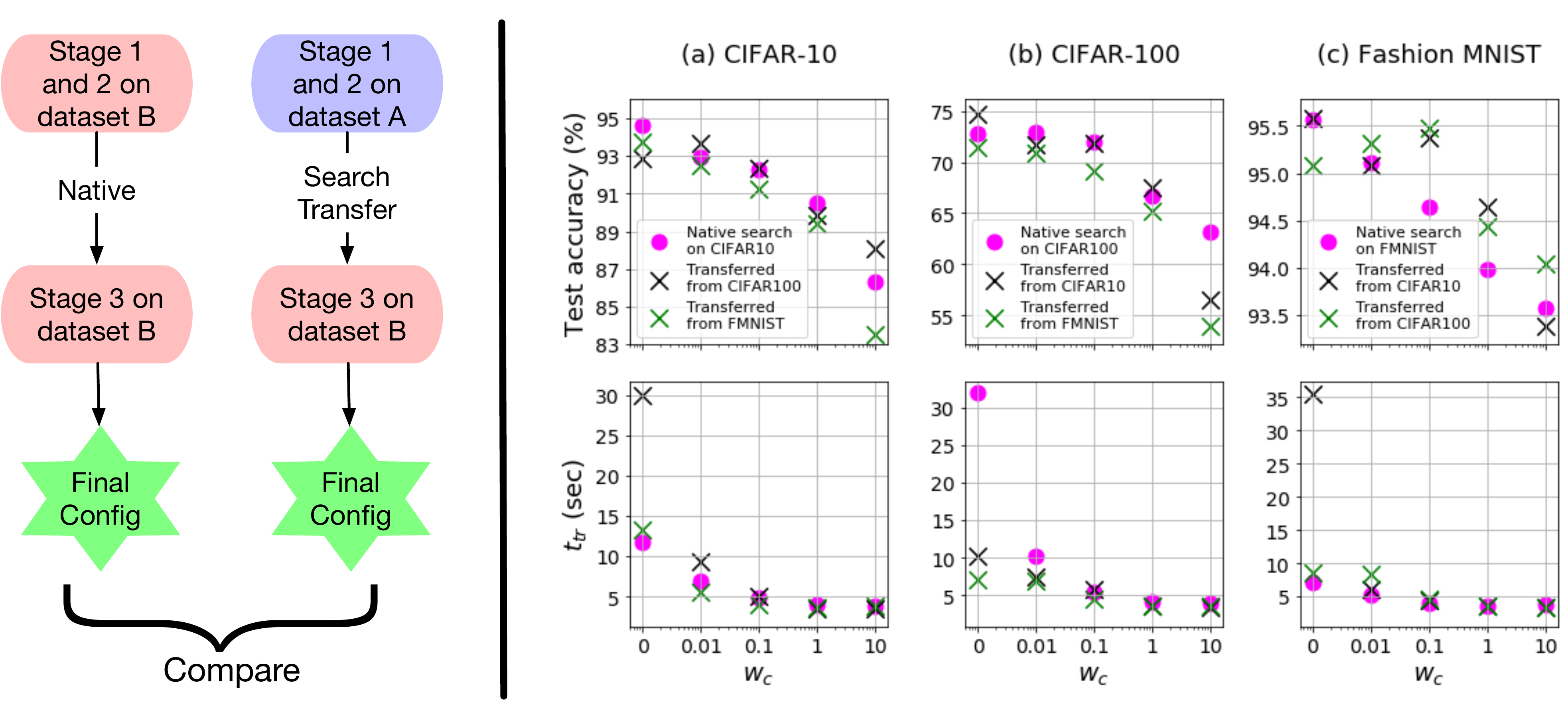}
    \caption{\emph{Left}: Process of search transfer -- comparing \acp{config} obtained from native search with those where Stage 3 is done on a dataset different from Stages 1 and 2. \emph{Right}: Results of \ac{CNN} search transfer to (a) CIFAR-10, (b) CIFAR-100, (c) \ac{FMNIST}. All datasets are augmented. Pink dots denote native search.}
    \label{fig-search_transfer}
\end{figure}

One goal of our search framework is to find models that are applicable to a wide variety of problems and datasets suited to different user requirements. To evaluate this aspect, we experimented on whether a \ac{NN} architecture found from searching through Stages 1 and 2 on dataset A can be applied to dataset B after searching for Stage 3 on it. In other words, how does transferring an architecture compare to `native' \acp{config}, \ie those searched for through all three stages on dataset B. This process is shown on the left in Fig. \ref{fig-search_transfer}. Note that we repeat Stage 3 of the search since it optimizes training hyperparameters such as weight decay, which are related to the capacity of the network to learn a new dataset. This is contrary to simply transferring the architecture as in \cite{Zoph2018_searchtransfer}.

We took the best \ac{CNN} architectures found from searches on CIFAR-10, CIFAR-100 and \ac{FMNIST} (as depicted in Fig. \ref{fig-cnn_results}) and transferred them to each other for Stage 3 searching. The results for test accuracy and $\ttr$ are shown on the right in Fig. \ref{fig-search_transfer}. We note that the architectures generally transfer well. In particular, transferring from \ac{FMNIST} (green crosses in subfigures (a) and (b)) results in slight performance degradation since those architectures have $N_p$ around 1M-2M, while some architectures found from native searches (pink dots) on CIFAR have $N_p>20$M. However, architectures transferred between CIFAR-10 and -100 often exceed native performance. Moreover, almost all the architectures transferred from CIFAR-100 (green crosses in subfigure (c)) exceed native performance on \ac{FMNIST}, which again is likely due to bigger $N_p$. We also note that $\ttr$ values remain very similar on transferring, except for the $w_c=0$ case where there is absolutely no time penalty.

\subsection{Greedy strategy}
\begin{figure}[!t]
    \centering
    \includegraphics[width=0.83\textwidth]{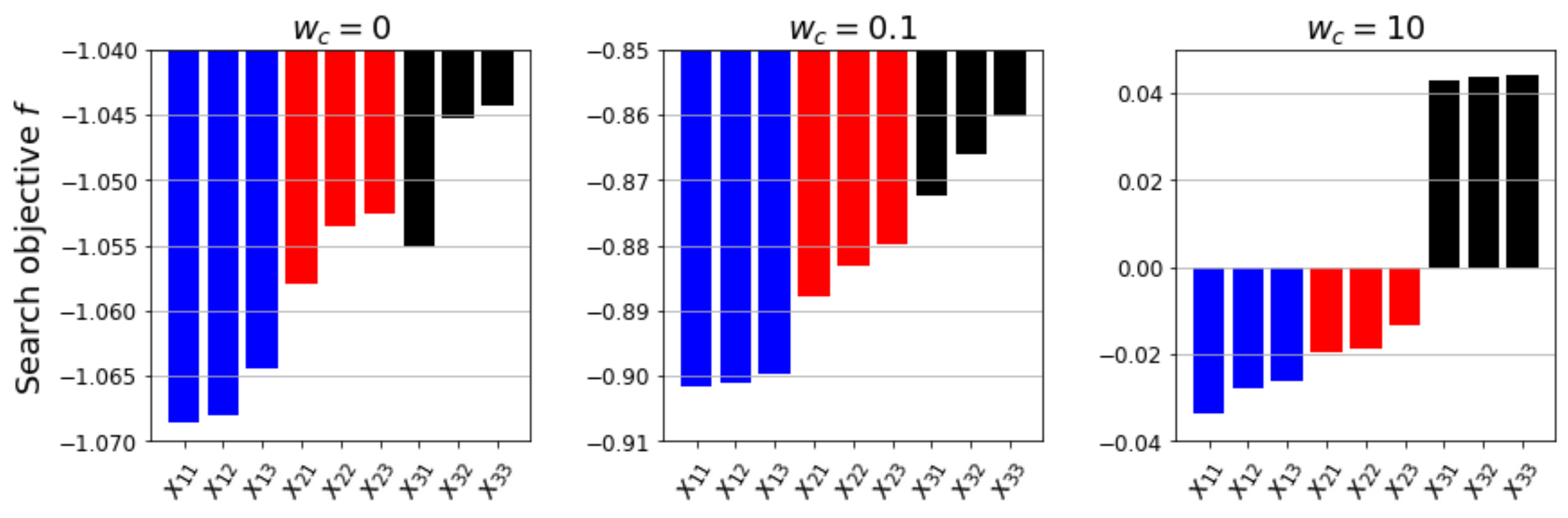}
    \caption{Search objective values (lower the better) for three best \acp{config} from Stage 1 (blue, red, black), optimized through Stages 2 and 3 and three best \acp{config} chosen for each in Stage 3. Results shown for different $w_c$ on CIFAR-10 unaugmented.}
    \label{fig-greedy}
\end{figure}

Our search methodology is greedy in the sense that it preserves only the best \ac{config} resulting in the minimum $f$ value from each stage and sub-stage. We also experimented with a non-greedy strategy. Instead of one, we picked the three best \acp{config} from Stage 1 -- $\left\{\bm{x}_1,\bm{x}_2,\bm{x}_3\right\}$, then ran separate grid searches on each of them to get three corresponding \acp{config} at the end of Stage 2, and finally picked the three best \acp{config} for each of their Stage 3 runs for a total of nine different \acp{config} -- $\left\{\bm{x}_{11},\bm{x}_{12},\bm{x}_{13},\bm{x}_{21},\cdots,\bm{x}_{33}\right\}$. Following a purely greedy approach would have resulted in only $\bm{x}_{11}$, while following a greedy approach for Stages 1 and 2 but not Stage 3 would have resulted in $\left\{\bm{x}_{11},\bm{x}_{12},\bm{x}_{13}\right\}$. We plotted the losses for each \ac{config} for five different values of $w_c$ on CIFAR-10 unaugmented (Fig. \ref{fig-greedy} shows three of these). In each case we found that following a purely greedy approach yielded best results, which justifies our choice for \ac{DnC}.

\subsection{Bayesian optimization vs random and grid search}
\begin{figure}[!t]
    \centering
    \includegraphics[width=0.83\textwidth]{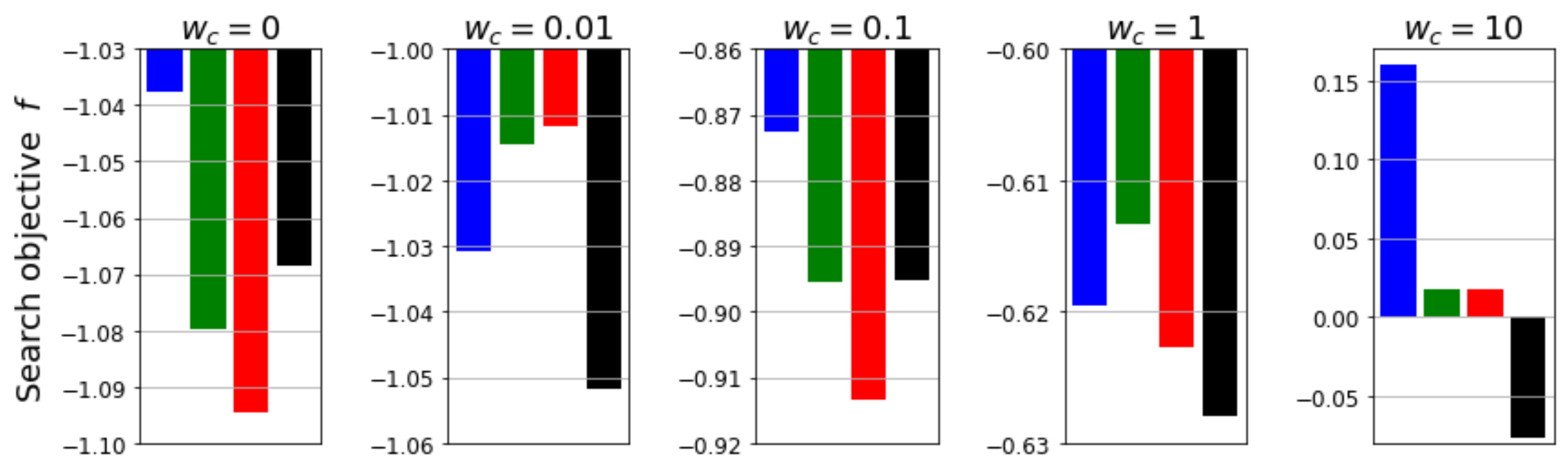}
    \caption{Search objective values (lower the better) for purely random search (30 samples, blue) vs purely grid search via Sobol sequencing (30 samples, green) vs balanced \ac{BO} (15 initial samples, 15 optimized samples, red) vs extreme \ac{BO} (1 initial sample, 29 optimized samples, black). Results shown for different $w_c$ on CIFAR-10 unaugmented.}
    \label{fig-rgbe}
\end{figure}

We use Sobol sequencing -- a space-filling method that selects points similar to grid search -- to select initial points from the search space and construct the \ac{BO} prior. We experimented on the usefulness of \ac{BO} by comparing the final search loss $f$ achieved by performing the Stage 1 and 3 searches in four different ways:
\begin{itemize}
    \item Random search: pick 30 prior points randomly, no optimization steps
    \item Grid search: pick 30 prior points via Sobol sequencing, no optimization steps
    \item Balanced \ac{BO} (\ac{DnC} default): pick 15 prior points via Sobol sequencing, 15 optimization steps
    \item Extreme \ac{BO}: pick 1 initial point, 29 optimization steps
\end{itemize}

The results in Fig. \ref{fig-rgbe} are for different $w_c$ on CIFAR-10. \ac{BO} outperforms random and grid search on each occasion. In particular, more optimization steps are beneficial for low complexity models, while the advantages of \ac{BO} are not significant for high performing models. We believe that this is due to the fact that many deep nets \cite{Komodakis2016_WRN} are fairly robust to training hyperparameter settings.

\section{Comparison to related work}\label{sec-comp}
\begin{table}[!t]
\renewcommand{\arraystretch}{1.1}
\caption{Comparison of features of AutoML frameworks}
\label{table-feature_comparison}
\centering
\begin{tabular}{|c|c|c|c|}
\hline
\multirow{2}{*}{Framework} & \multirow{2}{*}{Architecture search space} & Training & Adjust model \\
& & hyp search & complexity \\
\hline
\hline
Auto-Keras & Only pre-existing architectures & No & No \\
\hline
AutoGluon & Only pre-existing architectures & Yes & No \\
\hline
Auto-PyTorch & Customizable by user & Yes & No \\
\hline
Deep-n-Cheap & Customizable by user & Yes & Penalize $\ttr$, $N_p$ \\
\hline
\end{tabular}
\end{table}

Table \ref{table-feature_comparison} compares features of different AutoML frameworks. To the best of our knowledge, only \ac{DnC} allows the user to specifically penalize complexity of the resulting models
. This allows our framework to find models with performance comparable to other state-of-the-art methods, while significantly reducing the computational burden of training. This is shown in Table \ref{table-perf_comparison_cnn}, which compares the search process and metrics of the final model found for \acp{CNN} on CIFAR-10, and Table \ref{table-perf_comparison_mlp}, which does the same for \acp{MLP} on \ac{FMNIST} and RCV1 for \ac{DnC} and Auto-PyTorch only, since Auto-Keras and AutoGluon do not have explicit support for \acp{MLP} at the time of writing
.

Note that Auto-Keras and AutoGluon do not support explicitly obtaining the final model from the search, which is needed to perform separate inference on the test set after the search. As a result, in order to have a fair comparison, Tables \ref{table-perf_comparison_cnn} and \ref{table-perf_comparison_mlp} use metrics from the search process -- $\ttr$ is for the train set and the performance metric is best validation accuracy. These are reported for the best model found from each search. Auto-Keras and AutoGluon use fixed batch sizes across all models, however, Auto-PyTorch and \ac{DnC} also do a search over batch sizes. We have included batch size since it affects $\ttr$. Each \ac{config} for each search is run for the same number of epochs, as described in Sec. \ref{sec-expts}. The exception is Auto-PyTorch, where a key feature is variable number of epochs. 

\begin{table}[!t]
\renewcommand{\arraystretch}{1.1}
\caption{Comparing frameworks on \acp{CNN} for CIFAR-10 augmented on GPU}
\label{table-perf_comparison_cnn}
\centering
\begin{minipage}{\columnwidth}
\resizebox{\columnwidth}{!}{
\begin{tabular}{|c|c|c|c|c|c|c|}
\hline
\multirow{2}{*}{Framework} & Additional & Search cost & \multicolumn{4}{|c|}{Best model found from search} \\
\cline{4-7}
& settings & (GPU hrs) & Architecture & $\ttr$ (sec) & Batch size & Best val acc ($\%$) \\
\hline
\hline
Proxyless NAS\footnote{Since Proxyless NAS is a search methodology as opposed to an AutoML framework, we trained the final best model provided to us by the authors \cite{proxylessnas_pvtcomm}. This model was trained in \cite{Cai2018_proxylessnas} using stochastic depth and additional cutout augmentation \cite{proxylessnas_pvtcomm} -- yielding an impressive $97.92\%$ accuracy on their test set. The result shown here was obtained without cutout or stochastic depth, and the validation accuracy is reported to compare with the metrics available from Auto-Keras and AutoGluon. 
The primary point of including Proxyless NAS is to compare to a model with state-of-the-art accuracy that has been highly optimized for CIFAR-10.} & Proxyless-G & 96 & 537 \ac{conv} layers & 429 & 64 & 93.22 \\
\hline
Auto-Keras\footnote{Auto-Keras does not support image augmentation at the time of writing this paper \cite{autokeras_issue}, so we used results from the unaugmented dataset.} & Default run & 14.33 & Resnet-20 v2
& 33 & 32 & 74.89 \\
\hline
\multirow{2}{*}{AutoGluon} & Default run & \textbf{3} & Resnet-20 v1
& 37 & 64 & 88.6 \\
\cline{2-7}
& Extended run & 101 & Resnet-56 v1
& 46 & 64 & 91.22 \\
\hline
\multirow{2}{*}{Auto-Pytorch} & `tiny cs' & 6.17 & 30 \ac{conv} layers & 39 & 64 & 87.81 \\
\cline{2-7}
& `full cs' & 6.13 & 41 \ac{conv} layers & 31 & 106 & 86.37 \\
\hline
\multirow{3}{*}{Deep-n-Cheap} & $w_c=0$ & 29.17 & 14 \ac{conv} layers & 10 & 120 & \textbf{93.74} \\
\cline{2-7}
& $w_c=0.1$ & 19.23 & 8 \ac{conv} layers & 4 & 459 & 91.89 \\
\cline{2-7}
& $w_c=10$ & 16.23 & 4 \ac{conv} layers & \textbf{3} & 256 & 83.82 \\
\hline
\end{tabular}
}
\end{minipage}
\end{table}

\begin{table}[!t]
\renewcommand{\arraystretch}{1.1}
\caption{Comparing AutoML frameworks on \acp{MLP} for \ac{FMNIST} and RCV1 on GPU}
\label{table-perf_comparison_mlp}
\centering
\resizebox{\columnwidth}{!}{
\begin{tabular}{|c|c|c|c|c|c|c|c|}
\hline
\multirow{2}{*}{Framework} & Additional & Search cost & \multicolumn{5}{|c|}{Best model found from search} \\
\cline{4-8}
& settings & (GPU hrs) & MLP layers & $N_p$ & $\ttr$ (sec) & Batch size & Best val acc ($\%$)  \\
\hline
\hline
\multicolumn{8}{|c|}{Fashion MNIST} \\
\hline
\multirow{3}{*}{Auto-Pytorch} & `tiny cs' & 6.76 & 50 & 27.8M & 19.2 & 125 & \textbf{91} \\
\cline{2-8}
& `medium cs' & 5.53 & 20 & 3.5M & 8.3 & 184 & 90.52 \\
\cline{2-8}
& `full cs' & 6.63 & 12 & 122k & 5.4 & 173 & 90.61 \\
\hline
Deep-n-Cheap & $w_c=0$ & 0.52 & 3 & 263k & 0.4 & 272 & 90.24 \\
\cline{2-8}
(penalize $\ttr$) & $w_c=10$ & \textbf{0.3} & 1 & \textbf{7.9k} & \textbf{0.1} & 511 & 84.39 \\
\hline
Deep-n-Cheap & $w_c=0$ & 0.44 & 2 & 317k & 0.5 & 153 & 90.53 \\
\cline{2-8}
(penalize $N_p$) & $w_c=10$ & 0.4 & 1 & \textbf{7.9k} & 0.2 & 256 & 86.06 \\
\hline
\hline
\multicolumn{8}{|c|}{Reuters RCV1} \\
\hline
\multirow{2}{*}{Auto-Pytorch} & `tiny cs' & 7.22 & 38 & 19.7M & 39.6 & 125 & 88.91 \\
\cline{2-8}
& `medium cs' & 6.47 & 11 & 11.2M & 22.3 & 337 & 90.77 \\
\hline
Deep-n-Cheap & $w_c=0$ & 1.83 & 2 & 1.32M & 0.7 & 503 & \textbf{91.36} \\
\cline{2-8}
(penalize $\ttr$) & $w_c=1$ & \textbf{1.25} & 1 & \textbf{100k} & \textbf{0.4} & 512 & 90.34 \\
\hline
Deep-n-Cheap & $w_c=0$ & 2.22 & 2 & 1.6M & 0.6 & 512 & \textbf{91.36} \\
\cline{2-8}
(penalize $N_p$) & $w_c=1$ & 1.85 & 1 & \textbf{100k} & 5.54 & 33 & 90.4 \\
\hline
\end{tabular}
}
\end{table}


We note that for \acp{CNN}, \ac{DnC} results in both the fastest $\ttr$ and highest performance. The performance of Proxyless NAS is comparable, while taking 43X more time to train. This highlights one of our key features -- the ability to find models with performance comparable to state-of-the-art while massively reducing training complexity. 
The search cost is lowest for the default AutoGluon run, which only runs 3 \acp{config}. We also did an extended run for $\sim100$ models on AutoGluon to make it match with \ac{DnC} and Auto-Keras -- this results in the longest search time without significant performance gain.

For \acp{MLP}, \ac{DnC} has the fastest search times and lowest $\ttr$ and $N_p$ values -- this is a result of it searching over simpler models with few hidden layers. While Auto-PyTorch performs slightly better for the benchmark \ac{FMNIST}, our framework gives better performance for the more customized RCV1 dataset.

\section{Conclusion and Future Work}\label{sec-conc}
In this paper we introduced Deep-n-Cheap -- the first AutoML framework that specifically considers training complexity of the resulting models during searching. While our framework can be customized to search over any number of layers, it is interesting that we obtained competitive performance on various datasets using models significantly less deep than those obtained from other AutoML and search frameworks in literature. We also found that it is possible to transfer a family of architectures found using different $w_c$ values between different datasets without performance degradation. The framework uses Bayesian optimization and a three-stage greedy search process -- these were empirically demonstrated to be superior to other search methods and less greedy approaches.

\ac{DnC} currently supports classification using \acp{CNN} and \acp{MLP}. Our future plans are to extend to other types of networks such as recurrent and other applications of deep learning such as segmentation, which would also require expanding the set of hyperparameters searched over. The framework is open source and offers considerable customizability to the user. We hope that \ac{DnC} becomes widely used and provides efficient \ac{NN} design solutions to many users. The framework can be found at \url{https://github.com/souryadey/deep-n-cheap}.

\bibliographystyle{splncs04}
\bibliography{aaa_main.bib}

\clearpage

\section*{Appendix: Validity of our covariance kernel}
The validity of our covariance kernel can be proved as follows. We note that since $x_{ik}$ and $x_{jk}$ are scalars, $d$ in eq. \eqref{eq-distfn_ramp} is the Euclidean distance. It follows from the properties of the squared exponential kernel that $\sigma\left(x_{ik},x_{jk}\right)$ in eq. \eqref{eq-kernel_part} is a valid kernel function. So if a kernel matrix $\bm{\Sigma_k}$ were to be formed such that $\Sigma_{k_{ij}} = \sigma\left(x_{ik},x_{jk}\right)$, then $\bm{\Sigma_k}$ would be positive semi-definite. Writing eq. \eqref{eq-kernel_whole} in matrix form gives $\bm{\Sigma} = \sum_{k=1}^{K}{s_k\bm{\Sigma_k}}$. Since a convex combination of positive semi-definite matrices is also positive semi-definite, it follows that $\bm{\Sigma}$ is a valid covariance matrix.

\section*{Appendix: Ensembling}
One way to increase performance such as test accuracy is by having an ensemble of multiple networks vote on the test set. This comes at a complexity cost since multiple \acp{NN} need to be trained. We experimented on ensembling by taking the $n$ best networks from \ac{BO} in Stage 3 of our search. Note that this \emph{does not increase the search cost} as long as $n\leq n_1+n_2$. However, it does increase the effective number of parameters by a factor of exactly $n$ (since each of the $n$ best \acp{config} have the same architecture), and $\ttr$ by some indeterminate factor (since each of the $n$ best \acp{config} might have a different batch size).

\begin{figure}
    \centering
    \includegraphics[width=0.7\textwidth]{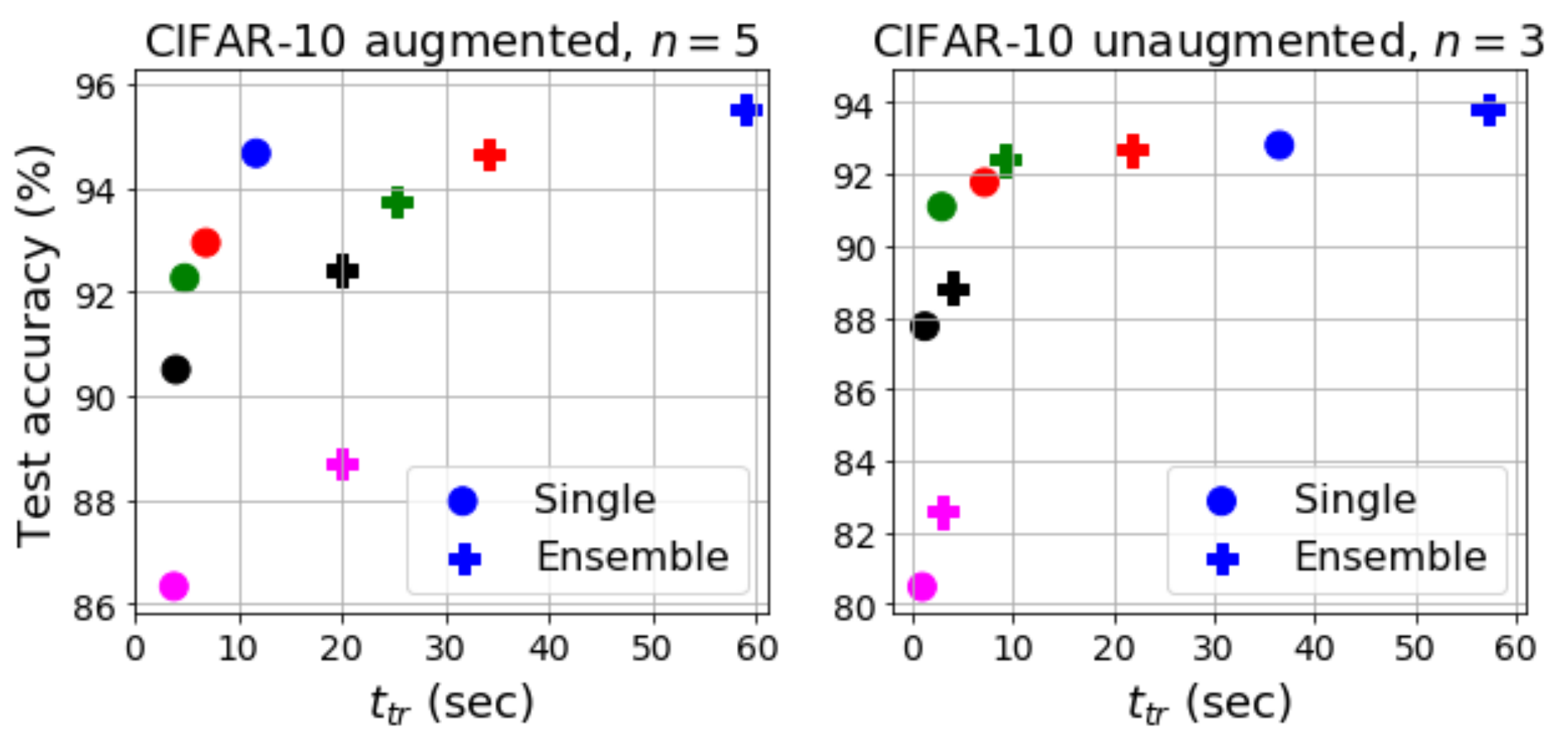}
    \caption{Performance-complexity tradeoff for single \acp{config} (circles) vs ensemble of \acp{config} (pluses) for $w_c = $ 0 (blue), 0.01 (red), 0.1 (green), 1 (black), 10 (pink). Results using ensemble of 5 for CIFAR-10 augmented, and 3 for CIFAR-10 unaugmented.}
    \label{fig-ensemble}
\end{figure}

We experimented on CIFAR-10 unaugmented using $n=3$ and augmented using $n=5$. The impact on the performance-complexity tradeoff is shown in Fig. \ref{fig-ensemble}. Note how the plus markers -- ensemble results -- have slightly better performance at the cost of significantly increased complexity as compared to the circles -- single results. However, we did not use ensembling in other experiments since the slight increases in accuracy do not usually justify the significant increases in $\ttr$.

\section*{Appendix: Changing hyperparameters of Bayesian Optimization}
The \ac{BO} process itself has several hyperparameters that can be customized by the user, or optimized using marginal likelihood or Markov chain Monte Carlo methods \cite{Swersky2013}. This section describes the default values we used. Expected improvement involves an exploration-exploitation tradeoff variable $\xi$. The recommended default is $\xi=0.01$ \cite{Brochu2010_BOtutorial}, however, we tried different values and empirically found $\xi=10^{-4}$ to work well. Secondly, $f$ is a noisy function since the computed values of network performance are noisy due to random initialization of weights and biases for each new state. Accordingly, and also considering numerical stability for the matrix inversions involved in \ac{BO}, our algorithm incorporates a noise term $\sigma_n^2$. We calculated its value from the variance in $f$ values as $\sigma_n^2 = 10^{-4}$, which worked well compared to other values we tried.

\section*{Appendix: Adaptation to various platforms}
While most deep \acp{NN} are run on GPUs, situations may arise where GPUs are not readily or freely available and it is desirable to run simpler experiments such as \ac{MLP} training on CPUs. \ac{DnC} can adapt its penalty metrics to any platform. For example, the \ac{FMNIST} results shown in Fig. \ref{fig-mlp_results} were on CPU, while Table \ref{table-perf_comparison_mlp} shows results on GPU (to do a fair comparison with other frameworks). As a result, the $\ttr$ values are an order of magnitude faster, while the performance is the same as expected.

\end{document}